\PassOptionsToPackage{numbers,sort&compress}{natbib}
\documentclass{article}

\usepackage[dblblindworkshop, final]{neurips_2026}
\workshoptitle{Agentic AI Benchmarks and Applications for Enterprise Tasks}

\usepackage[utf8]{inputenc} 
\usepackage[table]{xcolor}
\usepackage[T1]{fontenc}    
\usepackage{hyperref}       
\usepackage{url}            
\usepackage{booktabs}       
\usepackage{amsfonts}       
\usepackage{nicefrac}       
\usepackage{microtype}      
\usepackage{xcolor}         
\usepackage{graphicx}
\usepackage[most]{tcolorbox}
\usepackage{wrapfig}
\title{
Unified Agentic Video Editing Across Levels of Complexity and Creativity}

\author{
  \mbox{
    Surabhi S. Nath$^{1,2}$
    \qquad
    Kim Ferres$^{1,3}$
    \qquad
    Milan Petrović$^{1,3}$
    \qquad
    Lion Schulz$^{1}$
  }
  \\[0.2em]
  \mbox{
    $^{1}$Bertelsmann AI Hub
    \qquad
    $^{2}$Max Planck Institute for Biological Cybernetics
    \qquad
    $^{3}$scieneers GmbH
  }\\[0.2em]
  \hspace{-0.75cm}
  \texttt{surabhi.nath@tuebingen.mpg.de \hspace{0.7cm}
  lion.schulz@bertelsmann.nl}
}

\begin{document}

\maketitle

\vspace{-0.4cm}
\begin{abstract}
\vspace{-0.2cm}
Editing is a core component of video production, requiring creative planning and decisions under multiple constraints.
Here, we report methods for agentic tooling for automated video editing across three tasks varying in editorial goal, complexity and creativity, namely scene previews, video summaries and cinematic trailers. We evaluate the outputs and discuss implications for automation and agency. 


\end{abstract}

\vspace{-0.2cm}
\section{Introduction}
\vspace{-0.1cm}
The goosebumps we get from a great trailer boil down to great editing: thoughtful cuts and coherent storytelling. To achieve this, editors need to survey information, condense it, and make creative multimodal decisions, such as which storylines to tease or what music to use \citep{pearlman2025cutting}. This makes editing a natural setting for agentic AI \citep{acharya2025agentic,pati2025agentic,abou2025agentic}, combining planning, retrieval, with sequential decision making, tool use and evaluation of outputs \citep{xu2026agentic}. However, edits differ wildly in their editorial goal, complexity and creativity, something that efficient agentic enterprise editing solutions need to account for. Yet, all good editing shares the need for a strong and structured understanding of the source material. Here, we report experiments from a real-world production setting that show that one hierarchical data structure can support diverse editing tasks, from simple previews to cinematic trailers.

AI-assisted video editing has advanced in subtasks like searching and organising footage \citep{fu2026videostir,Leake2024ChunkyEdit,caballero2024automation}, supporting iteration and exploration \citep{huh2025videodiff,huh2026vidtune,yeh2026vidmento} or edits for specific content types \citep{argaw2024towards,pappachan2026quickcut,zhou2026autocut,dharmaratnakar2026generative}. More recent work explores agentic editing systems, decomposing complex tasks into sequences of smaller decisions \citep{Wang2024LAVE,SandovalCastaneda2025EditDuet,Zhao2026CutClaw,Wang2026BEAT,Zhou2026VideoAgent,Ding2025AgenticEditing,guo2026agentic}. However, real-world post-production and benchmarking remain challenging \citep{Cao2026AgenticVBench,argaw2022anatomy}, and comparisons of different tasks and orchestrations are lacking. Our work offers this unified perspective, and is relevant across communities: to the creative AI community for its integrated perspective, and to the agentic AI community for framing video editing as a frontier multimodal benchmark.
\vspace{0.25cm}

\begin{figure}[h!]
  \centering
  \includegraphics[width=\linewidth]{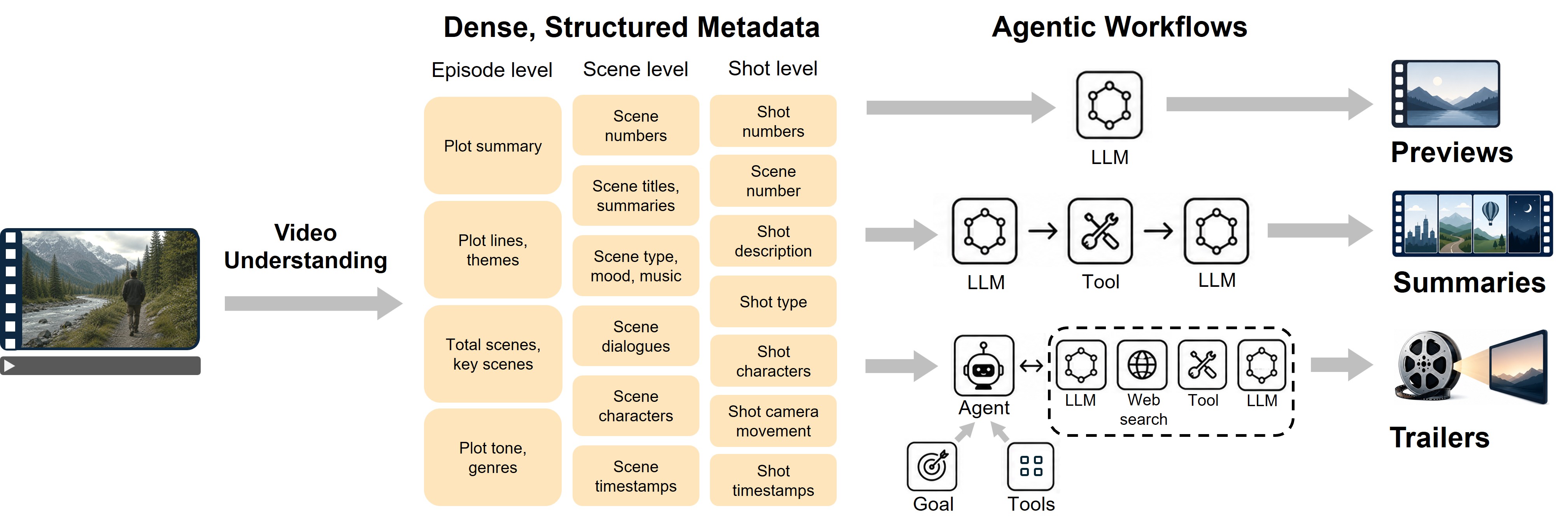}
  \caption{\textbf{Overview}: Tackling diverse editing tasks with a unified representation}
  \label{fig:overview}
\end{figure}

\newpage

\section{Methods and Results}
\vspace{-0.2cm}
\subsection{Task, system, data and evaluation overview}


While almost all editing tasks rely on retrieving, condensing, and re-assembling multimodal information, they can differ substantially in their editorial goal, complexity, and creative freedom. At the simpler end, previews involve selecting a single compelling scene under relatively few creative decisions, with the goal of hooking viewers. Video-to-video summaries introduce slightly higher structural complexity and creative freedom, while prioritising dense narrative information coverage. Cinematic trailers sit at the highest end of both complexity and creative freedom, combining many audiovisual components and involving creative re-narration to maximize hookiness while selectively withholding information. We here build and evaluate these three editing tasks that vary along these dimensions of \textbf{complexity}, \textbf{creative freedom}, \textbf{hookiness}, and \textbf{informativeness}.

We run these experiments on the first five episodes of an established and popular in-house TV production. 
Instead of raw video, agents interact with shared structured textual representations extracted from a proprietary video-understanding system (\ref{appendix1}). The system produces dense textual metadata at multiple temporal resolutions, comprising timestamped, hierarchically linked episode-, scene-, and shot-level descriptions and transcripts. 
This choice enables primarily text-driven editing workflows, letting agents reason over unified text input instead of multiple modalities, significantly reducing inference time and cost \citep{liu2025video,lin2025boosting}. \autoref{fig:overview} shows our end-to-end pipeline.

\subsection{Task 1: Preview generation}


We first consider a simple edit: selecting a representative contiguous scene, for uses such as auto-playing streaming carousels. The aim of such scenes is to entice a viewer to engage further with the material. Therefore, they should be \textit{high in hookiness}, while being spoiler-free, self-contained, dialogue-complete, and within a target duration. Though this involves constrained search, previews require relatively little explicit editing and offer limited creative autonomy. Leveraging the long-context and reasoning capabilities of modern LLMs, we apply a simple two-call workflow: the first assigns each scene a narrative role (setup, development, escalation, climax, resolution, breather or transition), and the second is prompted with all the specified editorial constraints to return a suitable clip (\ref{appendixpreviewprompts}).

To evaluate hookiness, we compare 45 generated previews against two baselines: an unguided LLM selection using the same context and duration but without the editorial instructions (ablated prompt), and an equal-duration clip sampled randomly (random pick). We use LLM-as-a-judge votes in a 2-AFC paradigm \citep{green1966signal} to measure clickability (which preview would you rather click on to watch the full video?) and quantify spoiler control by testing whether previews overlap fully with climax/resolution scenes.



\begin{figure}
    \centering
    \includegraphics[width=\linewidth]{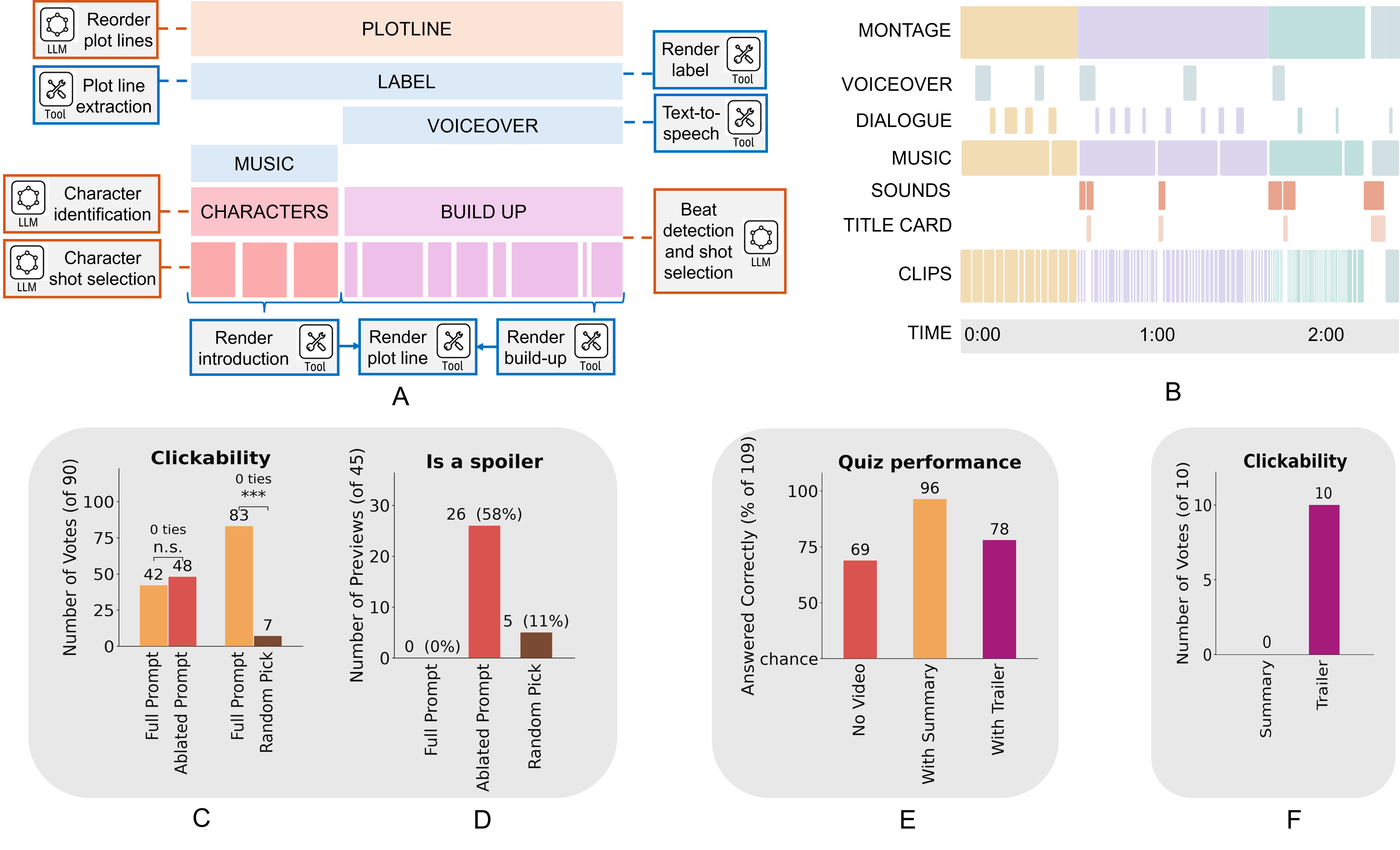}\vspace{-0.4cm}\caption{\textbf{Evaluations and output results}: (A) Summary output structure with associated LLM and tool calls. (B) Example trailer output structure. (C), (D) Preview evaluation for clickability and spoilers. (E) Quiz performance with different video contexts. (F) Trailer evaluation for clickability.}
    \label{figure2}
\end{figure}

Our results underscore the trade-off between clickability and spoiler avoidance. The ablated prompt received marginally more clickability votes than the full prompt (\autoref{figure2}C), but this difference is not significant and came at the expense of spoiling the story (\autoref{figure2}D). Compared to random pick, the previews from the full prompt were substantially more clickable (\textit{p} < 0.001, \autoref{figure2}C). Overall, the results suggest that explicit instructions help balance engagement and spoiler control.
We also evaluated understandability and edit quality which were superior for the full prompt (\ref{appendixpreviewsupp}).

\subsection{Task 2: Narrative video-to-video summarisation}

Simple preview scenes leave viewers guessing what comes next. In contrast, video-to-video summaries (VtVSs) are supposed to inform in as much detail as possible in a short amount of time. In production these edits are important for example in editorial, development, or sales functions when writers or clients need to get quick overviews. Video summaries also serve as the foundation for recaps that are played in the beginning of an episode. \textit{As such, VtVSs should have high informativeness.} VtVSs contain many intricacies that set them apart from simple text-to-text summaries. Particularly, if done well, they require reasoning over two modalities, audio and video, and arranging them in conjunction. For example, a naive editor might just concatenate the most relevant scenes. More sophisticated editors may exert some creative control to first introduce characters before building up plotlines.

To produce such more complex VtVS edits, we construct a structured multi-step workflow that decomposes edits into a sequence of intermediate decisions. First, an LLM re-orders the episode’s plotlines by narrative importance. Then, for each plotline, for an 8-second character-introduction segment, a second LLM identifies shots containing the relevant characters from our textual metadata, and a third LLM selects two representative shots per character. For the narrative buildup segment, the plotline description is converted into a voiceover using text-to-speech models. A fourth LLM call splits the description into narrative beats and selects 1–3 suitable shots for each beat. In the end, several tool calls render the resulting structured plan into the final summary.

As a result, we generate a layered edit combining clips, music, voiceover, and text on a common timeline (\autoref{figure2}A). Each plotline is marked by the textual label, a musical background introduces each character using two high-quality zoomed-in solo shots, followed by the buildup where a natural male voice narrates the plotline while the video layer unravels the story arc. Clips are selected such that their boundaries align well with the meaning conveyed by the corresponding voiceover beat. For example, for: "[X] gets promoted, and has a discussion with [Y]", the first couple clips show [X] getting promoted and the next shots show [X] talking to [Y], in sync with the voiceover narration.

To evaluate how informative these edits are, we create a benchmark comprising 22 multimodal multiple-choice questions (except one episode had 21 questions) probing knowledge about an episode (\ref{summaryeval}). We then compare the performance of a multimodal LLM on this quiz with the video summary loaded in its context, as opposed to without.
We found that LLMs with the video summary in their context answered 96\% of the questions correctly, compared to 69\% without video context (\autoref{figure2}E, \ref{summaryeval}). The relatively high no-context baseline compared to chance (25\%) reflects prior knowledge of the popular TV series and we would expect a lower floor for unreleased titles which we did not investigate here due to confidentiality. 
We also inspected the few errors made by the with-summary LLM and found that those questions contained finer details of key events not included in the summary edit. This highlights a trade-off between compression and detail retention typical for any summarisation task.



\vspace{-0.1cm}
\subsection{Task 3: Cinematic trailer editing}

For our final task, we stepped up the requirements significantly, considering a full cinematic, movie-style trailer. These edits contain many more components and decisions and push both editing complexity and creative freedom substantially further: montages, music, sound design, B-roll, title cards, climax teases, and stingers. Unlike previews and summaries, trailers follow less of a fixed stylistic structure or semantic objective: they may vary widely in pacing and audiovisual construction, and may even deliberately distort the story to create confusion and curiosity. \textit{They should therefore be high in hookiness, and low in informativeness}. 

In such a scenario, a hard-coded orchestration becomes too complicated and rigid. We therefore move from explicit workflow design to goal-directed agentic planning, directly specifying the desired creative outcome of generating a compelling trailer. 
We also let this system access the structured video metadata, and crucially, a collection of editing tools developed through the previous tasks such as scripts to find character shots, detect beats, separate audio and video, and align shots to voiceover. The agent is then responsible for deciding what steps to do, in what order, and which tools to invoke.  

We find that the combination of rich metadata and reusable editing tools enables the agent to develop a workflow of 24 steps, with 52 API calls, over 40 generations of text plans, voiceover and music, and 8 evaluation judgments (refer to \ref{appendixtrailer} for a further breakdown of the steps and calls involved). 

The resulting 2--3 minute trailers are well-planned and original, combining layers of dialogue, voiceover, music, sound design, title cards, silence beats, and over 100 clips on average (\autoref{figure2}B). Importantly, edit complexity is not only reflected in the number of components, but also in how they are temporally coordinated. For example, music changes mark the beginning of new montages, sound risers lead into title cards, shot pacing progressively increases through the trailer, with extremely fast cuts in the climactic third montage, and a silence beat  follows immediately to create contrast. In parallel to this audiovisual structuring, the story development is compelling, offering a substantially creative re-narration of the original plot.

Beyond qualitative evaluation, we set up two crucial comparisons vis-a-vis our video summaries. Namely, our trailers should be both \textit{more hooky} and \textit{less informative} than the summaries. To measure information, we repeat the quiz with trailer context to find that it performs better than no video context, but worse than summary video context (\autoref{figure2}E). We also conduct a matched 2-AFC clickability test between trailers and summaries, to find that the trailers are unanimously preferred (\autoref{figure2}F). Together, these results validate our intended distinction between a summary and trailer edit in terms of informativeness and hookiness. Besides this interpretable evaluation, we also received positive feedback on the trailer outputs from editors at our company.


\section{Discussion}
Our work investigated automated video editing using agentic AI for a range of production tasks with differing editorial goals---maximizing viewer engagement in previews, maximizing narrative coverage in video-to-video summaries, and balancing the two through a creative re-narration in cinematic trailers. We showed that a structured textual metadata backbone can support diverse agentic editing workflows, orchestrated depending on edit complexity (number of components and constraints) and creative freedom (determining how readily goals can be decomposed into well-defined subgoals).


Our work highlights how automating editing has important implications for human creative agency \citep{issak2026control, shin2026human, xu2025productive}. As workflows become agentic, editors may shift away from low-level, manual tasks like searching through video, towards specifying intent, constraints, and creative direction \citep{zhang2025exploring, zielinska2026creative}. 
Rather than automating the creative process, future agentic systems should act as support tools \citep{liu2026idesigngpt, das2026leveraging, nath2026designing} that help editors explore, refine, and get inspired by edits while retaining control over the final output.



We also show that video editing requires navigating competing objectives---between engagement and spoiler avoidance or compression and detail retention---making evaluation inherently multidimensional \citep{hopkins2026human,nath2025pencils} and challenging as creative autonomy grows: beyond verifiable properties, open-ended edits like trailers are shaped by the audience, enterprise needs, and subjective taste \citep{yao2025automatic}, which should be better captured in future benchmarks. More broadly, we position trailer generation as a challenging real-world testbed for how agents translate subjective, creative intent into coordinated actions.

Despite some success in producing high-quality edits, our work has limitations. Our case study is limited to five episodes from a single fictional, scripted TV show. Broader studies across productions, genres and content types are needed to test for workflow generality. Additionally, while expert editors within our enterprise qualitatively reviewed all outputs, a systematic, quantitative human assessment would help validate our findings beyond LLM-as-a-judge evaluation. Further, our agentic trailer generation had 24 steps with 52 API calls. This flags how agentic autonomy can introduce long execution chains, high costs, possible cascading of errors and high stochasticity \citep{mishra2025can,barke2026agentrx, esen2025risks}, risks we did not study systematically. Increased autonomy also introduces AI safety concerns \citep{chhabra2026agentic}. These range from enterprise concerns like  unintended disclosure of copyrighted material, misleading re-narrations, outputs violating brand regulations, to safety concerns. For example, trailers might enable agents to purposefully hide information or misdirect viewers, subtle forms of manipulation and deception \citep{carroll2023characterizing, alon2023dis, akbulut2026evaluating}. Future human-centred agentic systems should also focus on mitigating such concerns.



\newpage

\begin{ack}
Authors thank Mats Faulborn, Niket Kapoor, Richard Naab, Johannes Schmidt and Tom Wehmeyer for helpful feedback and contributions to the underlying metadata system.
SSN and LS are employees of Bertelsmann SE \& Co. KGaA and its affiliates. KF and MP are employees of scieneers GmbH, and were contracted to the Bertelsmann AI Hub for the duration of this project.
\end{ack}


\bibliographystyle{unsrtnat}

\bibliography{neurips26}


\newpage

\appendix

\section{Technical appendices and supplementary material}

\subsection{Video Understanding System}
\label{appendix1}

The in-house proprietary video understanding system combines vision-language models with classical machine learning analysis to extract rich, dense, structured textual metadata at hierarchical temporal resolutions. The episode-level metadata includes information on the plot summary, plotlines, themes, total number of scenes, key scenes, plot tone and genres. The scene-level metadata includes scene numbers, types, titles, summaries, moods, dialogues, characters and timestamps. The shot-level metadata includes shot numbers, corresponding scene numbers, shot type, narrative, description, characters, camera movement and timestamps. 

The pipeline is general-purpose and applies to videos of widely differing content types (for example scripted TV shows, reality TV, daily soaps, game shows, documentaries, movies), genres (for example action, romance, comedy, suspense/thriller, horror) and languages (for example English, German).

\newpage
\subsection{Details of Workflows}
\label{appendix2}

\subsubsection{Preview Prompts \scriptsize{[blue part not in ablated prompt]}}
\label{appendixpreviewprompts}
Generated with \texttt{gemini-3.1-pro-preview}: mean of 180k tokens, and median of \$0.43 per preview.

\noindent
\begin{minipage}[t]{0.485\textwidth}
\begin{tcolorbox}[
    colback=gray!8,
    colframe=gray!35,
    boxrule=0.5pt,
    arc=1.5pt]
\textbf{1. Narrative structure (once per video)\\}

You are a narrative structure analyst. Your task is to analyse the structural architecture of a video's narrative - not just what happens, but HOW the story is constructed.
\\\\
\textbf{Task}
\\\\
Analyse the narrative structure by:
1. Identifying the overall structural pattern
2. Locating major turning points (structural beats)
3. Defining act boundaries (if applicable)
4. Classifying each scene's structural role
\\\\
\textbf{Input}
\\\\
Plot Summary, Plotlines, Key scenes
\\\\
\textbf{Output}
\\\\
1. Pattern Identification: Identify which structural pattern best fits this narrative: three act, four act, five act, hero's journey, episodic, nonlinear, other.

2. Turning Points: Identify major structural beats. These are moments where the narrative shifts direction or reaches a significant structural milestone.

3. Act Divisions: If the narrative has clear act structure, define where each act begins and ends.

4. Scene Roles: Classify EVERY scene's structural role. Each scene gets exactly ONE role: Setup: Establishing characters, world, relationships, or situation, Development: Building plot, deepening relationships, adding complications, Escalation: Raising stakes, increasing tension or conflict, Climax: Peak conflict, confrontation, or major resolution moment, Resolution: Resolving conflicts, showing consequences, wrapping up, Breather: Pause in tension, emotional respite, character moment, Transition: Moving between locations, timeframes, or story threads
\\\\
Analyze the narrative structure now and provide the output in JSON format.

\end{tcolorbox}
\end{minipage}
\hfill
\begin{minipage}[t]{0.485\textwidth}
\begin{tcolorbox}[
    colback=gray!8,
    colframe=gray!35,
    boxrule=0.5pt,
    arc=1.5pt
]
\textbf{2. Generate preview (once per preview)\\}

You are an expert video editor. Choose a continuous time window from a video to use as candidate PREVIEW clips. \textcolor{blue}{Each preview should have broad, general appeal — a well-rounded, intriguing hook.
Each preview should also be appropriate for and representative of the video's
genre (genre, secondary genre).}
\\\\
\textcolor{blue}{\textbf{Requirements}}
\\\\
\textcolor{blue}{EACH window you pick MUST satisfy ALL of the following:
1. EXCITING \& SUSPENSEFUL — it should hook the viewer and make them want to
   watch the full video. Favour rising tension, intrigue, suspense, excitement,
   conflict, danger or a compelling character moment.\\
2. NOT A SPOILER — it must NOT reveal the outcome or ending. Specifically:
   do NOT include any part of a scene whose role is "climax" or "resolution", do NOT give away the twist, who wins/loses/dies, or how the central conflict resolves. Prefer buildup material: scenes with role "setup", "development", "escalation" or "breather".\\
3. STANDALONE \& SELF CONTAINED — the window must be intelligible without the rest of the show.\\
4. NEVER CUT A DIALOGUE — start time and end time must NOT fall inside any dialogue line. Every dialogue line that overlaps the window must be FULLY contained in it.}\\
5. MATCH THE TARGET DURATION — try to be as close to {duration} seconds.
\\\\
\textbf{Input}
\\\\
Plot summary, Key scenes, Scene Metadata
\\\\
\textbf{Output}
\\\\
Return start and end timestamps of the  candidate windows. Give a
reasoning in ONE short sentence (max ~20 words) explaining why it works as a preview.
\end{tcolorbox}
\end{minipage}

\subsubsection{Video-to-video summarisation}

\begin{wrapfigure}{r}{0.48\textwidth}
    \centering
    \vspace{-8pt}
    \includegraphics[width=1.05\linewidth]{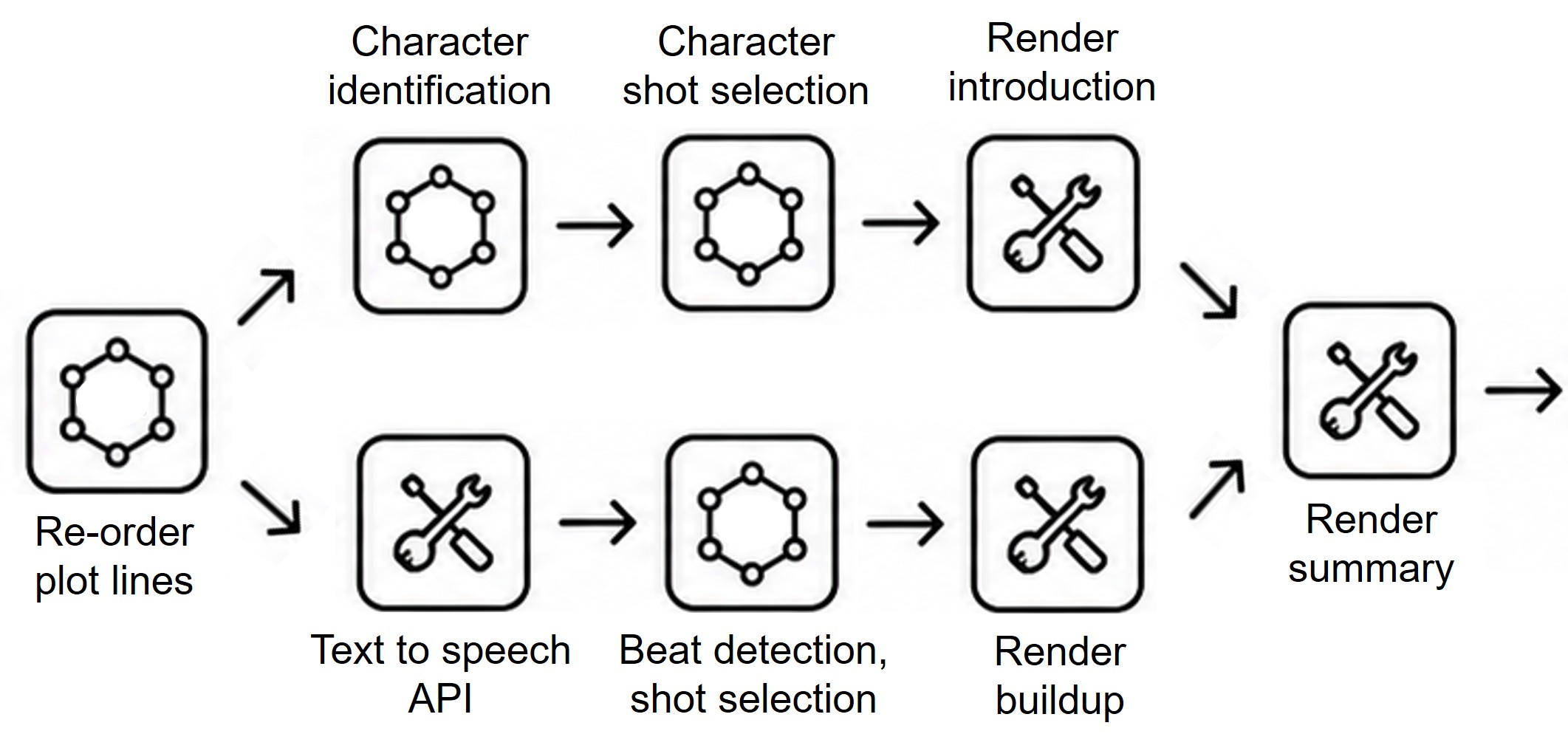}
    \caption{Summary Workflow}
    \label{figure3}
    \vspace{-8pt}
\end{wrapfigure}

The video-to-video summary workflow (\autoref{figure3}) received textual metadata of plotlines, scenes and shots as input and generated a 90-110 second output video organised as shown in (\autoref{figure2}A). The model used to generate the edits was \texttt{gemini-3.7-flash}. The voiceover used Google Cloud text-to-speech API, \texttt{Chirp 3 HD} in Charon voice. 

Three Gemini calls are made per plotline and one call in the beginning to order the plotlines, resulting in a total of 13 calls for an episode with 4 plotlines. Temperature was set between 0 and 0.2. The total number of input tokens was between 120-220k per summary costing a median of \$0.26 for \texttt{gemini-3.7-flash}.

\subsubsection{Trailer Generation}
\label{appendixtrailer}
For our agentic trailer generation, we provide \texttt{Claude Code 2.1.251} a set of high-level goals to make a trailer that is: creative, expert-level, cinematic, movie-like, that stuns a viewer, with full freedom for using various resources. The dense annotated metadata, along with the set of developed scripts/tools (including beat detection, word onset detection, text to speech, character shot identification, dialogue selection, picking shots relevant to text/audio, rendering tools, etc.) were linked, and Google API usage instructions were provided. A further breakdown of the workflow of 24 steps and 52 API calls is shown in \autoref{table1}. The VLM used was \texttt{gemini-3.1-pro-preview} and the median cost per trailer was $\$3.05$.

\begin{table}[h]
\scriptsize{}
\centering
\caption{Trailer generation pipeline describing all scripts, token usage and number of model calls.
\label{table1}
Token counts are with all calls summed. Music generation produces audio rather than tokens
and is billed per clip. Output tokens include thinking tokens, and for speech synthesis they are
audio tokens.}
\begin{tabular}{clllrrr}
\toprule
& Script & Stage & What it does & In tok. & Out tok. & Calls \\
\midrule
1  & \texttt{build\_index}    & Index          & Flatten annotations into shot + dialogue indices   & --      & --     & 0 \\
2  & \texttt{creative\_pass}  & Treatment      & Author the locked treatment from the full corpus   & 145,500 & 15,121 & 1 \\
3  & \texttt{validate\_spine} & Validation     & Check every quoted line against the dialogue index & --      & --     & 0 \\
4  & \texttt{lock\_spine}     & Lock           & Freeze the verified dialogue spine                 & --      & --     & 0 \\
5  & \texttt{audio\_qc}       & Audio QC       & Test each line's separability from the score       & 2,000   & 6,500  & 1 \\
6  & \texttt{alternates}      & Alternates     & Propose substitute lines per beat                  & 72,500  & 16,062 & 1 \\
7  & \texttt{audition\_lines} & Line audition  & Rate candidate clips by ear, in batches            & 6,800   & 30,752 & 4 \\
8  & \texttt{revise\_spine}   & Spine revision & Rebuild the spine from verified usable lines       & 4,000   & 11,658 & 1 \\
9  & \texttt{visual\_edl}     & Visual EDL     & Cut each movement against the locked spine         & 225,000 & 45,762 & 3 \\
10 & \texttt{build\_timeline} & Timeline       & Normalise the EDL to a frame-exact timeline        & --      & --     & 0 \\
11 & \texttt{contact\_sheets} & Contact sheets & Render sheets of every selected shot               & --      & --     & 0 \\
12 & \texttt{patch\_timeline} & Patch                & Apply the curated shot substitutions         & --     & --     & 0  \\
13 & \texttt{gen\_music}      & Music generation     & Generate cue variants per movement           & --     & --     & 13 \\
14 & \texttt{audition\_music} & Music audition       & Select cues by listening + energy analysis   & 13,600 & 7,956  & 1  \\
15 & \texttt{gen\_music\_2}   & Music generation & Generate cues for slots found weak           & --     & --     & 6  \\
16 & \texttt{vo\_audition}    & Voice casting        & Render candidate voices, pick by listening   & 6,900  & 9,017  & 13 \\
17 & \texttt{render\_vo}      & Narration            & Synthesise each narration line               & 1,000  & 768    & 5  \\
18 & \texttt{render\_clips}   & Picture render       & Cut, grade and stabilise each clip           & --     & --     & 0  \\
19 & \texttt{make\_cards}     & Title cards          & Render title and card elements               & --     & --     & 0  \\
20 & \texttt{build\_audio}    & Mix                  & Five-bus mix with side-chain ducking         & --     & --     & 0  \\
21 & \texttt{assemble}        & Assemble             & Join picture and audio                       & --     & --     & 0  \\
22 & \texttt{master}          & Master               & Two-pass loudness normalisation              & --     & --     & 0  \\
23 & \texttt{qc\_audio}       & Mix QC               & Measure narration intelligibility per window & 19,800 & 14,196 & 3  \\
24 & \texttt{final\_sheet}    & Final sheet          & Render the delivery contact sheet            & --     & --     & 0  \\
\midrule
\multicolumn{4}{l}{\textit{Total}} & 497,100 & 157,792 & 52 \\
\bottomrule
\end{tabular}
\end{table}

This pipeline is the output from an agent but is still a deterministic workflow and is a fixed chain of scripts. Future work would replace the fixed chain with a custom agentic harness exposing a chat interface, letting editors direct the cut conversationally, for example testing an alternative line, reordering a specific shot, swapping a title card, and seeing the consequence rendered immediately.

\begin{figure}
    \centering
    \includegraphics[width=0.8\linewidth]{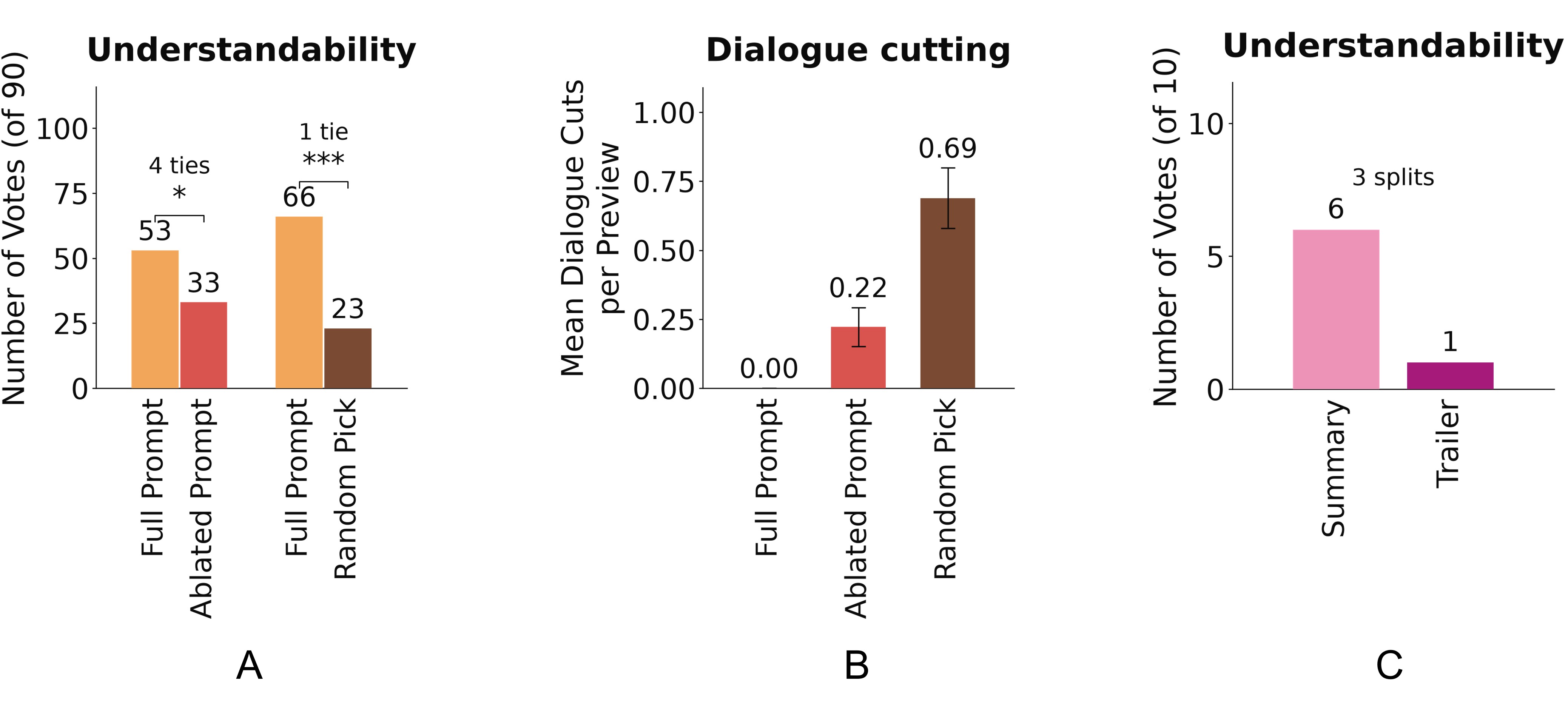}
    \caption{Supplementary 2-AFC results: (A) Understandability evaluation for previews, (B) Dialogue Cutting evaluation for previews, and (C) Understandability evaluation for summaries vs. trailers. Ties are cases where flipping the order of the options flipped the answer.}
    \label{figure4}
\end{figure}

\newpage
\subsection{Details of Evaluations}
\subsubsection{2-AFC}
\label{appendixpreviewsupp}

To test preview performance on \textit{hookiness}, we conducted a 2-AFC LLM-as-a-judge evaluation between previews generated by one of three methods: either using the full (second) prompt mentioned in \ref{appendixpreviewprompts} (full prompt), or that prompt without the blue parts (ablated prompt), or a fully random duration-matched pick from anywhere in the episode (random pick). In order to test the effect of the method, while taking into account the inherent model stochasticity, we repeated the experiment several times. We generated 3 previews per each of the 15 episode $\times$ duration combinations where episode can be \{1, 2, 3, 4, 5\} and duration can be \{15, 30, 45\} seconds, per method. This resulted in 45 previews per method, and we performed two sets of comparisons: full prompt vs. ablated prompt, and full prompt vs. random pick. Each judgement was given by two models: \texttt{gemini-3.1-pro-preview} and \texttt{gemini-3.7-flash} for both option orders, and the votes were pooled, yielding the total of 180 comparisons. Testing both option orders helps detect order effects: the cases where the judgment flipped with order switching were considered to be ties. Therefore, only 90 judgements are reported and the other 90 were made to determine model consistency and ties. The judges were asked two questions: clickability (which preview would you rather click on to watch the full video?, \autoref{figure2}C) and understandability (which preview is easier to understand on its own, without having seen the rest of the episode?). Spoiler control and dialogue cutting were measured descriptively as: does the preview overlap fully with a climax/resolution scene (Yes/No, \autoref{figure2}D), and how many dialogue lines are cut at the clip boundaries (0/1/2), respectively.

Besides effectively balancing clickability and spoiler avoidance (\autoref{figure2}C, D), we see that the full prompt also has superior performance with respect to both understandability: previews it generated were significantly more understandable as a standalone segment compared to the previews generated by other methods (\autoref{figure4}A), and dialogue cutting: no dialogue lines were cut in any preview generated by the full prompt (\autoref{figure4}B).

Further, we used this same paradigm to evaluate the hookiness of our generated trailers. Here, we compared trailers against video summaries, which serve as a class of edits with relatively low emphasis on hookiness. As expected, \autoref{figure2}F shows that trailers were consistently judged as more clickable than summaries. However, this comparison introduces a potential length confound: summaries are 90--110 seconds long on average, whereas trailers are 120--180 seconds. The longer trailers therefore expose judges to more audiovisual tokens, potentially biasing the comparison in their favour. 

We were able to largely rule out such a bias due to the results of LLM-as-a-judge votes on understandability for trailers vs. summaries. This serves as a good control as summaries are meant to be more informative and hence understandable than trailers that try to purposely twist the story or withhold information. In this case we find that despite containing more audiovisual tokens, trailers are rated as generally less understandable than summaries (\autoref{figure4}C).

\subsubsection{Quiz Benchmark}
\label{summaryeval}
To test for informativeness, we developed a multimodal quiz benchmark. Our quiz included 22 questions per episode across five bands of question types: 
\begin{enumerate}
    \item Premise: the episode’s main plotlines and conflicts [4 questions]. For example: \textit{What is the initial incident that brings [X] into contact with [Y]?} [4 options]
    \item Key events: the major events and resolutions of the plotlines [8 questions]. For example: \textit{What is the ultimate fate of [Z]?} [4 options]
    \item Plotline exclusion: the plotline that does not occur in the episode [1 question]: \textit{Which of the following storylines is NOT featured in this episode?} [4 options]
    \item Character recognition: the character photograph matching a given description [5 questions]. For example: \textit{Which of these characters is facing retirement?} [4 character options]
    \item Character role: the role of a given character in the episode [4 questions]. For example: \textit{Which two major life events is this man dealing with simultaneously in this episode?} [character image attached, 4 options]

\end{enumerate}

We then compare the performance of \texttt{gemini-2.5-flash} on this question set with no video context, the summary video context and with the trailer video context. Overall, the performance was best with summary video context, followed by trailer video context and lowest with no video context---which was still significantly higher than the random chance baseline (\autoref{figure2}E). 

\begin{figure}
    \centering
    \includegraphics[width=0.85\linewidth]{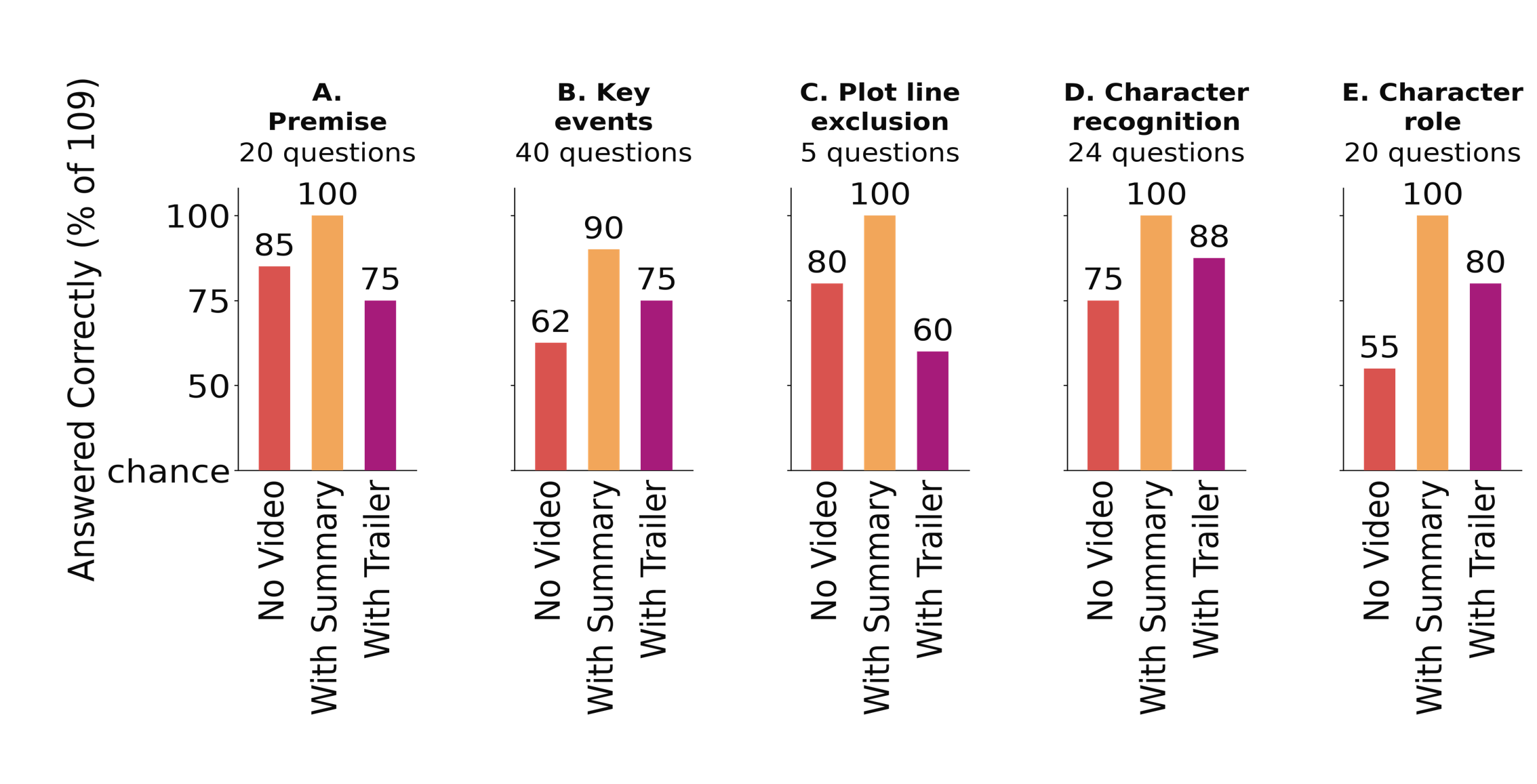}
    \caption{Performance on quiz benchmark per question band: comparison across no video context, summary video context and trailer video context.}
    \label{figure5}
\end{figure}

Here we report the performance per band to find that the performance with summary context is also consistently superior across bands (\autoref{figure5}). Interestingly however, performance with trailer context is worse than that with no video context in two out of five bands. This suggests that trailers can not only omit information, but actively distort or override the model's existing priors by presenting a potentially misleading account of the episode.

\end{document}